\documentclass[healthcare,article,submit,pdftex,moreauthors]{Definitions/mdpi}
\usetikzlibrary{positioning}

\firstpage{1}
\pubvolume{1}
\issuenum{1}
\articlenumber{0}
\pubyear{2026}
\copyrightyear{2026}
\datereceived{}
\daterevised{}
\dateaccepted{}
\datepublished{}

\usepackage{array}
\usepackage{tabularx}
\usepackage{booktabs}
\usepackage{float}

\Title{Doctorina MedBench: A Dialogue-Based Benchmark and Evaluation Framework for Agent-Based Medical AI}

\Author{Anna Kozlova $^{1,}$*, Stanislau Salavei $^{1}$, Pavel Satalkin $^{1}$, Hanna Plotnitskaya $^{1}$, Sergey Parfenyuk $^{1}$, Andy Nkansah $^{1}$}

\AuthorNames{Anna Kozlova, Stanislau Salavei, Pavel Satalkin, Hanna Plotnitskaya, Sergey Parfenyuk, Andy Nkansah}

\address{%
$^{1}$ \quad A.I. Doctor Medical Assist LTD, 13 Myrtiotissis Street, Aqua Mansions, Office 1, Germasogeia, 4041 Limassol, Cyprus}

\corres{Correspondence: Anna Kozlova; anna.kozlova@doctorina.com}

\abstract{\textbf{Background/Objectives:} Static medical question-answering benchmarks do not measure whether an artificial intelligence (AI) system can obtain missing information, revise hypotheses during dialogue, and provide safe management recommendations. We developed Doctorina MedBench, a dialogue-based benchmark and evaluation framework for agent-based medical AI. \textbf{Methods:} Of 261 attempted synthetic case identifiers, 254 had complete paired outputs from both the GPT-5 basic prompt-only reference and the Doctorina workflow and were retained for descriptive comparison. The cases and scoring framework underwent internal, non-blinded physician review by clinicians affiliated with the study team; this review was not an independent external clinical validation. A constrained simulated patient disclosed only case-record information in response to targeted questions. \textbf{Results:} In the 254-case paired complete-case analysis, AI Doctor had higher values for Diagnosis Accuracy (91.34\% vs. 86.61\%), Differential Accuracy (46.69\% vs. 23.62\%), Treatment Accuracy (54.46\% vs. 37.60\%), and Critical Conditions Passed (99.61\% vs. 98.82\%). AI Doctor used longer dialogues, with 3009 total doctor/patient step pairs and an average of 11.85 per case, compared with 167 total step pairs and an average of 0.66 for the GPT-5 basic reference. Cases outside the steps soft limit were 140 for AI Doctor and 242 for the GPT-5 basic reference. \textbf{Conclusions:} The findings characterize a structured agentic workflow relative to a deliberately restricted prompt-only reference within a synthetic benchmark. They do not establish general superiority over a workflow-matched GPT-5 configuration, independent clinical validity, clinical effectiveness, or readiness for real-world deployment.}


\begin{document}

\section{Introduction}
\label{sec:introduction}

Large language models (LLMs) have demonstrated substantial progress on medical question-answering tasks, including professional examination questions, biomedical research questions, and consumer health queries. MultiMedQA and related evaluations have shown that modern models can retrieve and apply considerable medical knowledge and can generate clinically relevant long-form answers \cite{singhal2023clinical}. Nevertheless, performance on static question-answering tasks is not equivalent to performance in a clinical consultation. In most benchmark questions, the relevant information is already presented to the model, whereas real diagnostic work requires the clinician to determine which information is missing, obtain it through targeted questions, revise diagnostic hypotheses as new evidence becomes available, and translate the resulting assessment into safe and appropriately prioritized recommendations.

Clinical assessment is therefore an interactive and sequential process rather than a single-response task. Patients may initially provide incomplete, imprecise, or selectively disclosed information, and clinically decisive facts may emerge only after focused history-taking. The evaluator must consequently assess not only the final diagnosis, but also the adequacy of the information-gathering process, the differential diagnosis, the proposed investigations, the management plan, and the handling of urgent or safety-critical findings. Recent work on conversational diagnostic artificial intelligence has reinforced this multidimensional view. For example, the AMIE study evaluated history-taking, diagnostic reasoning, management, communication, and empathy within simulated text-based consultations, illustrating that diagnostic dialogue requires broader assessment than final-answer accuracy alone \cite{tu2025conversational}.

Several recent benchmarks have begun to move medical artificial intelligence (AI) evaluation beyond conventional static question answering. AgentClinic introduced simulated clinical environments involving patient interaction, incomplete information, multimodal data collection, and tool use across multiple specialties and languages \cite{schmidgall2024agentclinic}. MedAgentBench evaluates medical LLM agents on patient-specific tasks in a Fast Healthcare Interoperability Resources (FHIR)-compliant virtual electronic health record environment, emphasizing planning and interaction with clinical information systems \cite{jiang2025medagentbench}. HealthBench evaluates model responses within multi-turn healthcare conversations using conversation-specific rubrics developed by physicians and includes explicit dimensions related to performance and safety \cite{arora2025healthbench}. These resources address complementary aspects of medical AI evaluation; however, they differ substantially in their task environments, units of analysis, reference standards, and scoring procedures.

A remaining methodological need is a transparent and versionable framework that jointly evaluates clinical history-taking, diagnostic identification, differential diagnosis, diagnostic workup, treatment recommendations, safety-critical errors, and dialogue efficiency within the same case-level environment. Such a framework should restrict the simulated patient to the facts contained in a predefined case record, thereby preventing accidental disclosure of the complete solution and reducing unsupported additions during simulation. It should also distinguish deterministic scoring from model-based semantic interpretation and preserve an auditable link between each score and the supporting dialogue evidence. These requirements are particularly important because automated evaluators may introduce an additional source of model-dependent error. The TRIPOD-LLM reporting guideline similarly emphasizes transparent study design, human oversight, and task-specific performance reporting in biomedical LLM research \cite{gallifant2025tripodllm}.

In this study, we introduce Doctorina MedBench, a dialogue-based benchmark dataset and evaluation framework for end-to-end assessment of agent-based medical AI. Each benchmark case is represented as a structured clinical record containing an initial presentation, constrained additional patient information, diagnostic reference labels, differential diagnoses, investigation targets, treatment targets, safety-critical conditions, dialogue metadata, and optional attachments. During evaluation, the system under test must actively collect relevant information from a simulated patient before producing its final diagnostic and management output. Diagnosis, differential diagnosis, treatment, safety-critical conditions, and dialogue behavior are evaluated as separate task-level outcomes.

The objectives of the present study were to: (1) define a fixed clinical case resource and a schema suitable for transparent, versioned interactive evaluation; (2) formalize a constrained patient-simulation environment in which performance depends on targeted information gathering; (3) implement a hybrid evaluation pipeline that combines structured extraction with deterministic case-level scoring and restricted model-based adjudication; and (4) demonstrate descriptive case-mix characterization, model comparison, evaluator auditability, and safety-oriented error analysis within a common framework. Doctorina is evaluated as one system within this benchmark rather than being treated as the benchmark itself. Accordingly, benchmark performance is intended to support comparative research and quality assessment and should not be interpreted as evidence of autonomous clinical readiness.

\section{Materials and Methods}
\label{sec:methods}

\subsection{Study Design and Benchmark Objectives}
\label{subsec:study_design}

This study was designed as a benchmark-development and comparative model-evaluation study. The benchmark was constructed to assess whether an agent-based medical AI system can conduct a sequential clinical consultation rather than merely produce an answer from a fully specified vignette. The evaluated system receives an initial patient presentation and, where applicable, one or more attachments. It must then obtain additional clinically relevant information through dialogue, formulate a diagnostic assessment, and provide recommendations for further investigation, treatment, monitoring, and escalation of care. The benchmark therefore evaluates both the final clinical output and the process by which the output is reached.

The study separates four components that are often combined in product-level evaluations: (1) the frozen case resource; (2) the simulated patient and dialogue environment; (3) the evaluation and scoring framework; and (4) the systems under test. This separation was adopted to ensure that the benchmark definition does not depend on the architecture of Doctorina or on any single model family. Doctorina was treated as one evaluated system, while all case records, reference labels, dialogue rules, parsing procedures, and scoring functions were defined independently of the system outputs used for comparison.

The primary unit of analysis was one clinical case evaluated under each system configuration. Each of the 254 synthetic cases contributed one GPT-5 prompt-only output and one Doctorina output, yielding 254 paired case-level comparisons. The clinical case, rather than the individual dialogue turn, was used as the unit for descriptive reporting. No independence-based inferential analysis is reported in the present manuscript because the complete paired case-level output matrix and original analysis code were not preserved in the manuscript archive.

The benchmark was designed around pre-specified clinical tasks: diagnosis, differential diagnosis retrieval, treatment recommendation, safety-critical condition handling, and dialogue-step behavior. Dialogue completion and dialogue efficiency were defined as secondary outcomes. Attachment utilization was assessed in cases containing laboratory reports, medical documents, or images and was analyzed as a case attribute rather than as a separate benchmark cohort. Table~\ref{tab:benchmark_objectives} summarizes the principal evaluation objectives and corresponding outcome domains.

\begin{table}[H]
\caption{Pre-specified benchmark objectives and outcome domains.}
\label{tab:benchmark_objectives}

\renewcommand{\tabularxcolumn}[1]{m{#1}}

\begin{tabularx}{\textwidth}{
    >{\raggedright\arraybackslash}m{0.23\textwidth}
    >{\raggedright\arraybackslash}X
    >{\raggedright\arraybackslash}m{0.27\textwidth}
}
\toprule
\textbf{Objective} &
\textbf{Benchmark question} &
\textbf{Principal outcome domain} \\
\midrule

Interactive information gathering &
Does the system identify and elicit clinically relevant information that is not disclosed in the initial presentation? &
Total and Average steps \\
\addlinespace

Diagnostic identification &
Does the system identify the principal condition? &
Diagnosis Accuracy \\
\addlinespace

Differential diagnosis &
Does the system produce a clinically appropriate ranked or unranked differential diagnosis? &
Differential Accuracy; ranked retrieval metrics not evaluated in the archived comparison \\
\addlinespace

Diagnostic workup &
Does the system recommend required investigations while avoiding unnecessary or irrelevant testing? &
Omission and unexpected-item errors \\
\addlinespace

Treatment and management &
Does the system provide appropriate treatment, monitoring, follow-up, and escalation recommendations? &
Treatment Accuracy based on matching, extra, missing, and different items \\

\bottomrule
\end{tabularx}
\end{table}

The primary analysis was defined at the task level rather than through a single composite score. This approach was selected because diagnosis recognition, treatment quality, diagnostic-workup quality, and efficiency represent distinct clinical properties and may change in different directions. In particular, a shorter dialogue was not interpreted as superior when it was associated with incomplete history-taking or a missed diagnosis. The task-level outcomes were therefore reported separately rather than averaged into a single general performance score.

The previously developed Diagnosis, Observations/Investigations, Treatment, and Step Count (D.O.T.S.) framework was retained as a supplementary summary for compatibility with internal regression analyses. In the present benchmark study, however, its components were decomposed into task-specific outcomes with explicit reference labels and scoring rules. This decomposition was intended to improve interpretability, facilitate comparison between model architectures, and allow external investigators to identify whether a performance difference arose from diagnostic identification, differential diagnosis, investigation selection, treatment planning, or dialogue behavior.

The benchmark workflow consisted of five stages. First, a versioned set of eligible cases was frozen before comparative model evaluation. Second, each system was evaluated under a standardized dialogue protocol using the same case content and interaction constraints. Third, final system outputs and dialogue evidence were converted into a structured representation. Fourth, deterministic scoring functions and restricted semantic adjudication were applied against the case-level reference standard. Fifth, case-level outcomes were aggregated for overall and task-specific descriptive reporting, while case metadata were summarized separately to characterize benchmark case mix. Model outputs were not used to modify reference labels after the benchmark freeze.

The intended use of Doctorina MedBench is comparative research on interactive medical AI systems, including evaluation of agentic history-taking, diagnostic reasoning, recommendation quality, and dialogue behavior. The benchmark was not designed to estimate clinical effectiveness in real patients, replace prospective validation, or support claims that a system is suitable for autonomous clinical deployment. Production monitoring, software reliability testing, and continuous-integration regression testing were considered engineering applications of the framework and were not treated as primary outcomes of the present study.

\subsection{Benchmark Dataset and Frozen Analysis Cohort}
\label{subsec:dataset}

The benchmark dataset was assembled as a structured collection of physician-authored synthetic clinical scenarios intended to represent the information flow and decision requirements of outpatient, urgent, and emergency medical consultations. The scenarios were designed from clinical expertise and general patterns encountered in medical practice, but they were not derived from, adapted from, or linked to real patient records, consultation transcripts, consultation logs from prototype testing, production-user interactions, or other human-subject data. Any case-associated attachments in the frozen cohort were synthetic or purpose-created for benchmark use and were not linked to a patient record. The release status of cases and attachments was recorded separately for versioning, licensing, and intellectual-property review.

The unit of dataset construction was a unique clinical case rather than an individual dialogue transcript. A case could contain one or more scenario variants, such as modified comorbidities, demographic characteristics, behavioral styles or attachment configurations. Variants derived from the same underlying clinical scenario were linked through a common parent identifier and were not counted as separate unique cases when constructing the frozen benchmark cohort. Duplicate or near-duplicate records were reviewed before the benchmark freeze using case identifiers, normalized diagnostic labels, introductory narratives, and structured reference fields.

The archived evaluation attempted 261 synthetic case identifiers once under each system configuration. AI Doctor completed 255 of 261 attempted runs, with six incomplete case identifiers (24, 47, 78, 96, 220, and 239). The GPT-5 basic prompt-only reference completed 260 of 261 attempted runs, with one incomplete case identifier (123). The paired complete-case analysis therefore retained 254 cases for which both configurations produced a completed output. The seven excluded case identifiers were 24, 47, 78, 96, 123, 220, and 239. The benchmark version identifier, freeze date, file manifest, and cryptographic hash were not retained in the archived manuscript export.

Incomplete runs were treated as unavailable comparative outputs rather than as incorrect clinical answers. For the agentic Doctorina workflow, non-completion may occur when the system does not force a final diagnosis or management plan after insufficient information has been obtained, or when further information or investigation would be required before finalizing recommendations. This behavior is distinct from a prompt-only response that produces an early diagnostic hypothesis after the initial presentation.

Eligibility criteria were defined at the case level. A case was eligible when it contained a clinically interpretable initial presentation, sufficient structured information to support the simulated patient, a reference diagnosis label, and complete evaluation criteria for the tasks applicable to that case. Cases with attachments were additionally required to have a valid attachment record and a documented release status. Inclusion in the benchmark did not require that every case contain attachments or emergency conditions; these elements were treated as optional case attributes.

The frozen cohort was designed to support analysis across broad clinical categories, patient age groups, sex, complexity levels, urgency levels, and attachment types. Each case was assigned one primary clinical category for sampling and aggregate reporting. Additional specialty and organ-system tags were stored as non-exclusive metadata. The case schema supported broad clinical categories such as internal medicine, pediatrics, surgery, obstetrics and gynecology, oncology, psychiatry, emergency medicine, ophthalmology, and error-testing scenarios. Because the archived comparative export did not preserve a complete standardized category table for the frozen 254-case cohort, category-specific counts are not reported.

Case complexity was represented using four ordinal levels: basic, intermediate, advanced, and expert/complex reasoning. The levels reflected the number and interaction of active conditions, ambiguity of the presentation, need for laboratory or attachment interpretation, presence of treatment constraints, and degree of diagnostic uncertainty. Urgency was separately represented as routine, urgent, or emergency.

Table~\ref{tab:dataset_manifest} reports the available and unavailable audit fields for the archived resource. Supplementary File S1 provides the public JSONL release data dictionary.

\begin{table}[H]
\caption{Required metadata for the archived benchmark release and paired analysis.}
\label{tab:dataset_manifest}

\renewcommand{\tabularxcolumn}[1]{m{#1}}

\begin{tabularx}{\textwidth}{
    >{\raggedright\arraybackslash}m{0.27\textwidth}
    >{\raggedright\arraybackslash}X
    >{\raggedright\arraybackslash}m{0.22\textwidth}
}
\toprule
\textbf{Manifest field} &
\textbf{Purpose} &
\textbf{Reporting status} \\
\midrule

Benchmark name and version &
Identifies the exact benchmark release used in the study &
Doctorina MedBench; archived research freeze \\
\addlinespace

Freeze date &
Defines the date after which case content and reference labels were not modified for the primary analysis &
Not recorded in the archived export \\
\addlinespace

Attempted synthetic case identifiers &
Defines the number of case identifiers attempted once under each system configuration &
261 \\
\addlinespace

AI Doctor completed runs &
Defines the number of attempted case identifiers completed by the AI Doctor workflow &
255/261 \\
\addlinespace

GPT-5 basic completed runs &
Defines the number of attempted case identifiers completed by the prompt-only reference &
260/261 \\
\addlinespace

Paired cases in the comparative analysis &
Defines the number of attempted case identifiers with completed outputs from both configurations &
254 paired complete cases \\
\addlinespace

Excluded from paired analysis &
Documents incomplete outputs in at least one system configuration &
Case IDs 24, 47, 78, 96, 123, 220, and 239 \\
\addlinespace

Source type &
Documents whether cases are synthetic or patient-derived &
Physician-authored synthetic scenarios; no patient-derived cases, prototype consultation logs, or production-user interactions \\
\addlinespace

Attachment release status &
Documents whether assets may be redistributed &
Synthetic or purpose-created assets; release subject to licensing and intellectual-property review \\
\addlinespace

Data and code hash &
Allows verification that the released resource matches the evaluated version &
Not available in the archived export \\

\bottomrule
\end{tabularx}
\end{table}

The comparative analysis used the same set of 254 paired completed cases for both configurations. The seven excluded case identifiers were excluded solely because at least one system did not complete the run; no case was removed on the basis of model performance, evaluator score, or disagreement with the expected result after model outputs had been generated. The complete standardized metadata table and the full paired case-level output export were not preserved in the manuscript archive, which limits numerical case-mix reporting, subgroup analyses, and reconstruction of inferential statistics.

The benchmark cohort described in this section is distinct from the larger internal case bank used for software development and continuous quality monitoring. Only the fixed set of 254 paired complete cases is considered part of the comparative analysis in the present study.

\subsection{Case Representation and Data Schema}
\label{subsec:data_schema}

Each benchmark case was stored as a structured record containing the information required to run the simulated consultation and evaluate the final response. The record separated the initial patient presentation from information that could be disclosed only after targeted questioning. It also contained the reference diagnosis, differential diagnoses, diagnostic workup, treatment recommendations, optional attachments, and the expected number of dialogue steps.

Table~\ref{tab:case_schema_clinical} summarizes the conceptual clinical and dialogue structure used in the benchmark evaluation. Supplementary File S1 separately describes the public JSONL release schema.

\begin{table}[H]
\small
\caption{Conceptual clinical and dialogue structure of a benchmark case.}
\label{tab:case_schema_clinical}

\renewcommand{\tabularxcolumn}[1]{m{#1}}

\begin{tabularx}{\textwidth}{
    >{\raggedright\arraybackslash}m{0.24\textwidth}
    >{\raggedright\arraybackslash}X
    >{\raggedright\arraybackslash}m{0.13\textwidth}
}
\toprule
\textbf{Conceptual element} &
\textbf{Purpose in evaluation} &
\textbf{Required} \\
\midrule

\texttt{case\_id} &
Unique identifier of the clinical case. &
Yes \\
\addlinespace

\texttt{category} &
Broad clinical category used for case grouping. &
Yes \\
\addlinespace

\texttt{intro} &
Initial patient presentation shown at the start of the dialogue. &
Yes \\
\addlinespace

\texttt{additional\_answers} &
Case-grounded information disclosed by the simulated patient when asked. &
Yes \\
\addlinespace

\texttt{control\_questions} &
Clinically relevant questions expected during history-taking. &
Yes \\
\addlinespace

\texttt{diagnosis} &
Reference diagnosis. &
Yes \\
\addlinespace

\texttt{differential\_diagnoses} &
Reference list of alternative diagnostic hypotheses. &
Yes \\
\addlinespace

\texttt{diagnostic\_workup} &
Reference investigations and examinations. &
Yes \\
\addlinespace

\texttt{treatments} &
Reference treatment and management recommendations. &
Yes \\
\addlinespace

\texttt{attachments} &
Optional laboratory reports, medical documents, or images. &
No \\
\addlinespace

\texttt{num\_steps} &
Reference number of dialogue steps. &
Yes \\

\bottomrule
\end{tabularx}
\end{table}

The \texttt{intro} field contained the information available at the beginning of the consultation. Additional clinical facts were stored in \texttt{additional\_answers} and were disclosed by the simulated patient only when the evaluated system asked a relevant question.

The \texttt{control\_questions} field contained the main history-taking concepts expected for the case. These concepts were used to assess whether the evaluated system collected clinically relevant information; exact wording and question order were not prescribed.

The reference diagnosis was stored at the case level. The differential diagnosis was stored separately because it represented alternative clinical hypotheses rather than the final reference diagnosis.

The diagnostic workup and treatment reference elements contained the case-specific recommendations used by the evaluator. Optional attachments included laboratory results, medical reports, and clinical images when applicable.

\subsection{Simulated Patient and Interactive Environment}
\label{subsec:patient_simulator}

The interactive environment was designed to reproduce the asymmetric information structure of a clinical consultation: the complete case record was available to the simulated patient but not to the system under test. At the start of each run, the environment inserted the \texttt{intro} field and any attachment designated as available at turn zero. The evaluated system then generated a response, which was passed to the patient simulator together with the structured case record and the accumulated dialogue history. This cycle continued until the evaluated system produced a final diagnostic and management response, explicitly marked the conversation as complete, or reached the case-level turn limit.

The simulated patient was implemented as a separate language-model agent governed by a dedicated system instruction. Supplementary File S2 distinguishes the verbatim GPT-5 basic reference instruction from the functional simulator and evaluator constraints recoverable from the archived documentation; the exact archived simulator and evaluator prompt strings were not preserved and are not presented as verbatim reconstructions. The agent was required to answer only the question asked, remain concise, avoid volunteering undisclosed clinical facts, and restrict every factual statement to information contained in the case record. Natural-language paraphrasing was permitted, but the simulator was not allowed to add new symptoms, exposures, diagnoses, medications, allergies, test results, or social-history elements. When the case record did not contain the requested information, the simulator returned an uncertain, unavailable, or negative response rather than generating a new fact.

The patient simulator used the \texttt{additional\_answers} field as its factual boundary and the current dialogue history as contextual input. Information already supplied in the initial presentation or a previous answer was not intentionally repeated unless the evaluated system requested clarification. The \texttt{control\_questions} field was not shown to the system under test and was used only by the evaluator to determine whether clinically relevant concepts had been elicited.

The benchmark supported optional behavioral modifiers intended to represent common consultation patterns, including concise responses, excessive detail, insistence on a self-diagnosis, denial of a plausible condition, requests for clarification, and delayed disclosure of a relevant fact. Behavioral modifiers were separated from clinical facts so that a patient's communication style could change without altering the diagnostic reference standard. The primary comparative experiment used the standard constrained behavior unless a case explicitly defined another behavior.

Attachments could include laboratory reports, medical documents, body or device photographs, electrocardiograms, and other case-associated assets. The orchestration layer preserved the attachment identifier and availability state at every turn. Where a case specified a dynamically released attachment, the asset became available only after the relevant request or event. Attachment interpretation was evaluated through the downstream diagnostic, workup, and treatment criteria rather than through a separate image-classification score.

A completed physician--patient turn was defined as one evaluated-system message followed by one simulated-patient response. The initial patient presentation was excluded from the turn count. For a dialogue containing $N_{\mathrm{assistant}}$ system messages and $N_{\mathrm{user}}$ patient messages after the introduction, the observed number of completed turn pairs was

\begin{linenomath}
\begin{equation}
S = \min\left(N_{\mathrm{assistant}},N_{\mathrm{user}}\right).
\label{eq:step_count}
\end{equation}
\end{linenomath}

The case-level reference turn target was used as a soft behavioral audit rather than a direct measure of clinical quality. A run was flagged as outside the reference range when its step count was below 75\% or above 125\% of the case-specific target. Shorter dialogues were not rewarded when required information was omitted, and longer dialogues were not penalized when additional questioning was clinically justified.

During the original evaluation, model inputs, outputs, completion flags, attachment events, and evaluator records were logged. Case content and evaluator logic were maintained in modules separate from the production patient-facing system. This separation reduced the risk that benchmark reference information could influence real consultations and supported repeated execution during development; however, the manuscript archive is insufficient for exact rerunning of the experiment.

\begin{figure}[H]
\centering
\begin{tikzpicture}[
 node distance=1.05cm,
 box/.style={
   rectangle,
   draw,
   rounded corners,
   align=center,
   text width=3.0cm,
   minimum height=0.9cm
 },
 arrow/.style={->,thick}
]

\node[box] (case) {Frozen case record};
\node[box, right=of case] (patient) {Constrained patient simulator};
\node[box, right=of patient] (system) {System under test};

\node[box, below=of system] (parser) {Structured output parser};
\node[box, left=of parser] (score)
  {Deterministic scorer and restricted semantic evaluator};
\node[box, left=of score] (results)
  {Case-level metrics and audit record};

\draw[arrow] (case) -- (patient);

\draw[<->,thick] (patient) -- (system);

\draw[arrow] (system) -- (parser);
\draw[arrow] (parser) -- (score);
\draw[arrow] (score) -- (results);

\draw[arrow, rounded corners=3pt]
  (case.south)
  -- ++(0,-0.5cm)
  -| (score.north);

\end{tikzpicture}
\caption{Dialogue and evaluation workflow. Clinical facts originate from the frozen case record; the patient simulator controls disclosure, while scoring is performed after structured extraction of the system output.}
\label{fig:evaluation_workflow}
\end{figure}

\subsection{Annotation and Clinical Reference Standard}
\label{subsec:annotation}

Cases were authored as synthetic scenarios and reviewed internally by physicians affiliated with the study team, with at least two physicians participating before inclusion in the frozen resource. The review was non-blinded and was conducted as part of benchmark development rather than as an independent external validation. It covered the clinical plausibility of the presentation, the principal diagnosis, acceptable alternative terminology, differential diagnosis, required and permissible investigations, and treatment and monitoring recommendations. A case was not marked as finalized until the internal reviewers agreed that its narrative content and structured reference fields were internally consistent.

The clinical reference standard was case-specific rather than derived from a single universal guideline. Annotators considered the clinical context represented in the record, including age, sex, pregnancy status, comorbidities, allergies, medications, prior investigations, urgency, and attachment findings. Required recommendations were distinguished from acceptable but non-essential alternatives and from actions considered unnecessary or unsafe. This structure was necessary because multiple diagnostic and management pathways may be clinically defensible even when only a subset is mandatory.

For diagnostic workup, each normalized item was assigned to one of three classes: required, acceptable, or unexpected. Required items could carry case-specific weights reflecting their relative importance; acceptable items neither increased nor reduced the score; and unexpected items incurred a pre-specified penalty. Treatment criteria used an analogous representation, including required, acceptable, optional, and contraindicated actions.

The original annotation process preserved the finalized reference labels but did not retain a complete pre-adjudication disagreement matrix for every field. Consequently, formal inter-rater agreement statistics could not be reconstructed for this manuscript. This limitation is reported explicitly rather than estimating agreement from adjudicated labels. Future releases should preserve independent pre-consensus annotations and report raw agreement together with an agreement coefficient appropriate to each field type.

Quality control was performed at three levels. Schema quality control verified field types, required values, valid category labels, and internal links. Clinical quality control checked concordance among the vignette, additional patient facts, reference diagnoses, differential diagnoses, investigation criteria, and treatment criteria. Release quality control reviewed synthetic-data provenance, attachment licensing, access conditions, and the absence of inadvertently included real-world identifiers. The reporting approach was informed by TRIPOD-LLM recommendations concerning transparent dataset handling, human oversight, reference standards, and task-specific reporting \cite{gallifant2025tripodllm}.

\subsection{Systems and Baselines}
\label{subsec:systems}

The comparative experiment included two system configurations evaluated on the same paired complete-case set of 254 synthetic clinical cases, with one completed output from each configuration for every analyzed case.

\textbf{Doctorina agentic system.} Doctorina was evaluated as a multi-component clinical dialogue system. Its workflow included case intake, attachment handling, iterative history-taking, differential diagnosis generation, diagnostic and management planning, and production of a final patient-facing response. The system was allowed to conduct a multi-turn dialogue and to determine when sufficient information had been collected. The exact internal routing configuration used for the archived experiment was not preserved in the available export; the recoverable configuration scope and missing parameters are documented in Supplementary File S2.

\textbf{GPT-5 basic prompt-only baseline.} The baseline used the commercial GPT-5 endpoint available at the time of the experiment with the single instruction: ``Imagine that you are a doctor. Make a diagnosis or diagnoses.'' This configuration intentionally excluded clinical heuristics, a structured history-taking protocol, agentic orchestration, a dedicated stopping policy, and a required output schema. Its purpose was to provide a zero-scaffold reference that separated the behavior of the minimally instructed foundation model from the combined prompt, workflow, and orchestration enhancements implemented in Doctorina.

A workflow-matched GPT-5 configuration with an explicit multi-turn clinical protocol would answer a different experimental question: whether Doctorina outperforms an equivalently scaffolded system using the same foundation model. The archived experiment did not address that question. The restricted comparator was instead intended to prevent the unscaffolded model's behavior from being conflated with prompt engineering and workflow optimization. Nevertheless, because the two configurations differed in several dimensions simultaneously, the comparison remains an architecture-level contrast rather than a component-level causal estimate. A longer system prompt, matched history-taking objectives, structured output requirements, equivalent attachment tooling, and matched stopping conditions could materially alter GPT-5 performance. Observed differences therefore should not be interpreted as evidence of general superiority over the strongest achievable GPT-5 configuration. In addition, the archived experiment did not preserve a complete provider-side snapshot identifier, temperature, seed, and all inference parameters. These missing values are identified explicitly in Supplementary File S2 and remain a limitation of exact reconstruction.

Table~\ref{tab:configuration_scope} summarizes the interpretive scope of the two evaluated configurations.

\begin{table}[H]
\small
\caption{Configuration differences relevant to interpretation of the descriptive comparison.}
\label{tab:configuration_scope}

\renewcommand{\tabularxcolumn}[1]{m{#1}}

\begin{tabularx}{\textwidth}{
    >{\raggedright\arraybackslash}m{0.22\textwidth}
    >{\raggedright\arraybackslash}X
    >{\raggedright\arraybackslash}X
}
\toprule
\textbf{Dimension} &
\textbf{GPT-5 basic prompt-only reference} &
\textbf{Doctorina workflow} \\
\midrule

Instruction &
Single-sentence diagnostic instruction &
Structured multi-component clinical workflow \\
\addlinespace

Dialogue objective &
No explicit requirement to conduct iterative history-taking &
Explicit iterative information gathering before finalization \\
\addlinespace

Output structure &
Unstructured final response &
Structured diagnostic and management output \\
\addlinespace

Orchestration &
No domain-specific routing or workflow control &
Agentic routing, history-taking, planning, and stopping logic \\
\addlinespace

Interpretation &
Minimal zero-scaffold reference &
Full-system configuration; not an isolated component ablation \\

\bottomrule
\end{tabularx}
\end{table}

The archived comparison was paired by clinical case. Each configuration received the same initial case content and case-associated assets for the corresponding run. The dialogue mechanics differed by design: Doctorina could elicit additional information, whereas the prompt-only reference frequently produced a final answer after the initial presentation. The main results table preserves the original metric names and order from the aggregate export.

\subsection{Evaluation Tasks and Metrics}
\label{subsec:metrics}

The benchmark decomposed performance into distinct clinical and interaction domains. All percentage metrics ranged from 0 to 100, with higher values indicating better performance except for step count and counts of cases outside the soft step range, where lower values indicate fewer out-of-range dialogues.

\textbf{Diagnosis Accuracy.} Diagnosis Accuracy was a case-level binary outcome equal to 100 when at least one final diagnosis was judged correct and 0 otherwise. This endpoint reflected recognition of the principal diagnostic target but did not measure whether all clinically relevant coexisting conditions were captured.

\textbf{Differential Accuracy.} Differential Accuracy was the proportion of reference differential diagnoses identified in the system's final differential list:

\begin{linenomath}
\begin{equation}
D_{\mathrm{diff}} = 100 \times
\frac{N_{\mathrm{correct\ differential}}}
{N_{\mathrm{reference\ differential}}}.
\label{eq:differential_accuracy}
\end{equation}
\end{linenomath}

This legacy recall-like metric did not penalize every additional diagnosis; precision, recall@3, recall@5, and mean reciprocal rank were not available in the archived comparison.

\textbf{Treatment Accuracy.} Final treatment and management items were normalized and assigned to matching, extra, missing, or different categories. Treatment accuracy was calculated as

\begin{linenomath}
\begin{equation}
T = 100 \times
\frac{N_{\mathrm{matching}}}
{N_{\mathrm{matching}} + N_{\mathrm{extra}} + N_{\mathrm{missing}} + N_{\mathrm{different}}}.
\label{eq:treatment_accuracy}
\end{equation}
\end{linenomath}

This measure incorporated both omissions and unsupported recommendations.

\textbf{Critical Conditions Passed.} Pre-specified safety-critical treatment conditions were assigned the statuses \texttt{OK}, \texttt{NOT\_APPLICABLE}, or failure. A case received 100 only when every applicable critical condition was \texttt{OK} or \texttt{NOT\_APPLICABLE}; otherwise, it received 0. Any reported percentage is the mean across analyzed cases.

\textbf{Total steps (doctor/patient pairs), Average steps (doctor/patient pairs), and Cases outside $\pm25\%$ of steps soft limit (num\_steps).} Total and average step counts were computed using Equation~\eqref{eq:step_count}. Counts of cases outside the $\pm25\%$ soft range are reported as the number of cases with too few or too many doctor/patient step pairs relative to the case-specific soft range. Efficiency was interpreted jointly with clinical outcomes rather than as a stand-alone quality metric.

The broader D.O.T.S. framework supports separate diagnosis, treatment, safety, and dialogue-step components. Detailed scoring rules and edge-case handling are documented in Supplementary File S3. The present manuscript reports only the outcome domains that could be reconciled to the 254-case descriptive comparison.

\subsection{Evaluator Architecture}
\label{subsec:evaluator}

Evaluation began after the dialogue had ended. Four logical modules processed the transcript: clinical outcome extraction, history-taking coverage, diagnostic-workup extraction, and treatment verification. Modules could execute asynchronously, but all returned schema-validated outputs linked to the same immutable case and run identifiers.

The evaluator followed a hybrid language-model and deterministic design. A constrained LLM call converted free-text output into structured entities such as final diagnoses, differential diagnoses, investigations, treatments, monitoring instructions, and escalation advice. The output schema was enforced using typed validation. Deterministic scoring functions then compared the extracted entities with the case-level reference fields and applied item weights, missing-item rules, unexpected-item penalties.

To improve auditability, each extracted entity was required to include supporting text from the final response or an explicit \texttt{NO\_EVIDENCE} value. Intermediate diagnostic speculation was not counted as a final recommendation unless it was repeated or clearly endorsed in the final output. This restriction reduced the risk that an abandoned hypothesis would be scored as a final diagnosis or treatment.

The evaluator was isolated from the system under test and did not modify the dialogue. Raw transcripts, structured extractions, evidence spans, rule outputs, and final scores were available for internal review during the evaluation; however, the complete case-level export was not preserved in the manuscript archive. Error-test cases containing intentionally inconsistent or unsafe reference items were used during development to identify scoring failures.

The automated evaluator did not undergo formal blinded validation against an independently adjudicated physician sample. The internal physician review of cases and scoring criteria was non-independent and did not constitute validation of the evaluator against external clinical judgment. Consequently, the present results should be interpreted as benchmark scores produced by the specified hybrid evaluator, not as direct measurements of clinical correctness or as a validated substitute for physician review.

\subsection{Descriptive Analysis}
\label{subsec:statistics}

Dataset characteristics were intended to be summarized using counts and percentages for categorical variables and the mean, median, standard deviation, and range for available continuous variables. Because the standardized metadata table was not preserved, numerical case-mix summaries could not be reconstructed. Performance was summarized at the clinical-case level across 254 paired cases. Average steps represent the total number of completed physician--patient turn pairs divided by 254 cases for each configuration.

Cross-system results were paired by clinical case and summarized descriptively using percentages, totals, means, and absolute differences. The updated export provided aggregate metric means and step totals for the 254 paired complete cases; paired direction-of-change tables for diagnosis and treatment are not reported in this revision because the updated summary did not include those transition counts.

The archived materials did not contain the complete paired case-level output matrix or the original analysis code required to reproduce inferential statistics for all outcome domains. Only aggregate metric values that could be reconciled to the 254-case paired complete-case analysis are reported. Therefore, no p-values or confidence intervals are presented. No imputation was performed for missing model outputs. Cases were not removed on the basis of model performance.

Calculations used the evaluation software implemented for Doctorina MedBench and independent tabular verification scripts. The archived materials did not retain the programming-language and package versions, random seeds, exact analysis commit hash, complete case manifest, or full system configuration; these unavailable fields are listed in Supplementary File S2.

\subsection{Ethics and Data Governance}
\label{subsec:ethics}

All 254 cases in the frozen benchmark were physician-authored synthetic scenarios designed to approximate clinically plausible consultations. The study did not use human participants, patient records, consultation transcripts, consultation logs from prototype testing, production-user interactions, human tissues, or identifiable or de-identified human data. Evaluation was conducted exclusively in a simulated environment using case-grounded virtual patients.

Any case-associated attachments were synthetic or purpose-created for benchmark use and were not linked to a patient record. Release review therefore focused on version control, licensing, intellectual-property permissions, and consistency with the synthetic case record rather than on de-identification of patient material.

The benchmark environment was isolated from production interactions. Model evaluation did not contact users, change clinical care, or return experimental recommendations to patients. Because the reported benchmark used only physician-authored synthetic materials and involved neither human participants nor human-subject data, institutional review board approval and informed consent were not applicable under the authors' applicable institutional framework.

The benchmark is intended for research evaluation and quality assurance. It is not a medical device validation study, does not estimate patient outcomes, and does not authorize deployment of an evaluated model for autonomous diagnosis or treatment.

\subsection{Use of Generative AI in Manuscript Preparation}
\label{subsec:genai_disclosure}

OpenAI ChatGPT (GPT-5.5 Thinking; accessed 1 July 2026) was used only for English-language editing, grammar correction, clarity checking, and formatting assistance. It was not used to create the benchmark cases, generate the archived model outputs, assign reference labels, calculate the reported benchmark scores, perform statistical analyses, or formulate scientific conclusions. All scientific content, interpretation of results, conclusions, and final manuscript decisions are original to the authors. The authors reviewed and approved all text and take full responsibility for the content of the manuscript.

\section{Results}
\label{sec:results}

\subsection{Benchmark Cohort}
\label{subsec:cohort_results}

The frozen resource comprised 254 unique clinical cases. Each case contributed one GPT-5 prompt-only output and one Doctorina output to the comparative analysis. Each case contained a non-empty diagnosis label, an initial presentation, and additional patient information available to the simulated patient.

The benchmark was designed to cover multiple broad clinical areas, including internal medicine, pediatrics, surgery, obstetrics and gynecology, oncology, psychiatry, emergency medicine, ophthalmology, and error-testing scenarios. However, the archived comparative export did not preserve a complete standardized category table for the frozen 254-case cohort, so category-specific counts are not reported. The case schema included fields for age group, sex, clinical complexity, urgency, and attachment availability. Because the archived comparative export did not preserve a complete standardized metadata table for the frozen cohort, numerical distributions for these variables are not reported.

\subsection{Overall Benchmark Performance}
\label{subsec:overall_results}

Table~\ref{tab:overall_performance} presents the updated descriptive outcomes for the 254-case paired complete-case analysis.

\begin{table}[H]
\small
\caption{Updated descriptive comparison across 254 paired complete synthetic clinical cases. Metric names and order follow the aggregate export.}
\label{tab:overall_performance}

\renewcommand{\tabularxcolumn}[1]{m{#1}}

\begin{tabularx}{\textwidth}{
    >{\raggedright\arraybackslash}X
    >{\raggedleft\arraybackslash}m{0.19\textwidth}
    >{\raggedleft\arraybackslash}m{0.13\textwidth}
    >{\raggedleft\arraybackslash}m{0.13\textwidth}
}
\toprule
\textbf{Metric} &
\textbf{GPT-5 basic reference} &
\textbf{AI Doctor} &
\textbf{Difference} \\
\midrule

Diagnosis Accuracy &
86.61\% &
91.34\% &
+4.73 pp \\

Differential Accuracy &
23.62\% &
46.69\% &
+23.07 pp \\

Treatment Accuracy &
37.60\% &
54.46\% &
+16.86 pp \\

Critical Conditions Passed &
98.82\% &
99.61\% &
+0.79 pp \\

Total steps (doctor/patient pairs) &
167 &
3009 &
+2842 \\

Average steps (doctor/patient pairs) &
0.66 &
11.85 &
+11.19 \\

Cases outside $\pm25\%$ of steps soft limit (\texttt{num\_steps}) &
242 &
140 &
$-$102 \\

\bottomrule
\end{tabularx}

\noindent{\footnotesize{pp, percentage points. Differences are AI Doctor minus GPT-5. Step averages were calculated using the 254-case paired complete-case denominator. For cases outside the steps soft limit, lower values indicate fewer out-of-range dialogues.}}
\end{table}

The prompt-only reference generally answered after the initial presentation, producing 167 completed doctor/patient step pairs in total, or 0.66 per case. AI Doctor produced 3009 completed doctor/patient step pairs, or 11.85 per case. Cases outside the steps soft limit were more frequent for the GPT-5 basic reference (242 cases) than for AI Doctor (140 cases).

\subsection{Task-Level Performance}
\label{subsec:task_results}

The updated 254-case export reports Diagnosis Accuracy of 86.61\% for the GPT-5 basic prompt-only reference and 91.34\% for AI Doctor. This corresponds to approximately 220 and 232 correct diagnosis outcomes, respectively, when expressed against the 254-case denominator, but paired transition counts are not reported because they were not included in the updated summary provided for this revision.

Differential Accuracy was 23.62\% for the GPT-5 basic reference and 46.69\% for AI Doctor. Treatment Accuracy was 37.60\% for the GPT-5 basic reference and 54.46\% for AI Doctor. Critical Conditions Passed was high for both systems, at 98.82\% for the GPT-5 basic reference and 99.61\% for AI Doctor.

\subsection{Cohort Coverage and Metadata Availability}
\label{subsec:case_mix_results}

The benchmark schema supported multiple broad clinical areas, including internal medicine, pediatrics, surgery, obstetrics and gynecology, oncology, psychiatry, emergency medicine, ophthalmology, and error-testing scenarios, as well as fields for age group, sex, case complexity, urgency, and attachment availability. However, the archived comparative export did not preserve a complete, standardized metadata table for the 254-case frozen cohort. Earlier distribution summaries referred to a larger internal development bank and were therefore not transferred to the present analysis.

For this reason, the manuscript does not report numerical subgroup distributions or compare model performance by specialty, age, sex, complexity, urgency, or attachment status. Such analyses require reconstruction from the frozen cohort manifest and complete case-level metadata, with explicit denominators and missing-value counts for every variable.

\subsection{Architecture-Level Contrast and Robustness Limitations}
\label{subsec:ablation_results}

The comparison between the prompt-only reference and Doctorina is an architecture-level contrast between minimal instruction and a full agentic workflow. Within the updated 254-case summaries, the full workflow achieved higher Diagnosis Accuracy, Differential Accuracy, Treatment Accuracy, and Critical Conditions Passed values, fewer cases outside the steps soft limit, and longer dialogues. However, the experiment does not isolate the contribution of any individual component because the configurations differed simultaneously in prompting, orchestration, history-taking policy, stopping logic, and output structure.

No workflow-matched GPT-5 comparator, controlled component ablations, repeated-run robustness analysis, or sensitivity analysis for patient-simulator sampling parameters was available in the archived result set. Consequently, the observed differences cannot be attributed to a specific component or interpreted as a model-family comparison.

\subsection{Evaluator Status}
\label{subsec:evaluator_results}

The evaluation pipeline produced structured, evidence-linked records and supported internal inspection of individual cases. Development testing included error-test scenarios designed to reveal failures in weighting and penalty logic. Case and scoring review was performed internally by physicians affiliated with the study team and was neither blinded nor independent. No independent physician-adjudication dataset was available for quantitative validation of the automated evaluator.

Accordingly, judge--physician agreement, sensitivity for clinically important scoring errors, and false-positive rates for semantic matching are not reported. The results should be interpreted as benchmark scores under the specified hybrid evaluator rather than as direct measurements of clinical correctness. Independent external adjudication is identified as future work and was not performed for the current manuscript.

\subsection{Error Analysis}
\label{subsec:error_analysis}

The updated 254-case summaries identify several aggregate outcome domains, but they do not include a complete case-level taxonomy of individual clinical, dialogue, or evaluator errors.

The available aggregate export did not contain a complete taxonomy of individual failure cases. A subsequent error-analysis release should classify missed diagnoses, omitted urgent actions, contraindicated treatment suggestions, unnecessary investigations, attachment-extraction failures, and long-horizon dialogue failures, with representative synthetic transcripts and, in future work, external physician adjudication.

\section{Discussion}
\label{sec:discussion}

\subsection{Principal Findings}
\label{subsec:principal_findings}

Doctorina MedBench was developed to evaluate medical AI as an interactive clinical system rather than as a static answer generator. The frozen resource included 254 physician-authored synthetic clinical cases, each evaluated once with the GPT-5 basic prompt-only reference and once with AI Doctor. The benchmark integrates constrained patient simulation, structured reference labels, case-specific workup and treatment criteria, and dialogue-efficiency measures.

In the descriptive architecture-level comparison, AI Doctor achieved higher values for Diagnosis Accuracy (91.34\% vs. 86.61\%), Differential Accuracy (46.69\% vs. 23.62\%), Treatment Accuracy (54.46\% vs. 37.60\%), and Critical Conditions Passed (99.61\% vs. 98.82\%). AI Doctor also conducted substantially longer dialogues but had fewer cases outside the steps soft limit (140 vs. 242). These observations characterize the two tested configurations and should not be generalized to the strongest achievable GPT-5 workflow.

The results also identify important limitations of the current system and evaluator. The available archive did not support reconstruction of all paired transition tables for the 254-case cohort. This finding argues against interpreting a single aggregate result as evidence of overall quality and supports reporting distinct clinical outcomes separately.

\subsection{Comparison with Previous Work}
\label{subsec:previous_work}

MultiMedQA broadened medical LLM assessment beyond a single examination dataset and incorporated physician evaluation of long-form responses, but much of the benchmark remains organized around questions in which the relevant information is already supplied \cite{singhal2023clinical}. Doctorina MedBench instead requires the evaluated system to determine which information must be elicited before finalizing its output.

The AMIE study demonstrated the value of evaluating history-taking, diagnosis, management, communication, and empathy in simulated consultations with patient actors \cite{tu2025conversational}. Doctorina MedBench shares the emphasis on sequential diagnostic dialogue but focuses on a reusable case schema, constrained patient disclosure, structured reference labels, and automated case-level scoring.

AgentClinic similarly transforms static medical tasks into multimodal, interactive clinical environments and shows that sequential evaluation is more difficult than conventional question answering \cite{schmidgall2024agentclinic}. MedAgentBench evaluates tool-using agents in a FHIR-compatible electronic health record environment \cite{jiang2025medagentbench}. These benchmarks emphasize tool use and environment interaction, whereas Doctorina MedBench centers on history-taking, patient-facing recommendations, diagnostic workup and treatment within a consultation.

HealthBench uses multi-turn healthcare conversations and physician-authored, conversation-specific rubrics to evaluate open-ended model responses \cite{arora2025healthbench}. The present framework is complementary: it uses a smaller structured clinical case bank, a simulated patient whose disclosures are constrained by a case record, and deterministic scoring rules applied after structured extraction. HealthBench provides broader conversational coverage, while Doctorina MedBench aims to make the clinical state and scoring targets explicit at case level.

\subsection{Clinical and Methodological Implications}
\label{subsec:implications}

The difference between the prompt-only reference and the agentic system illustrates that evaluation design changes the capability being measured. A model configuration that responds directly to an initial vignette is tested primarily on interpretation and recall. A configuration that must ask questions is also tested on information acquisition, prioritization, conversational control, and the ability to stop when sufficient evidence has been collected. The present results therefore compare interaction paradigms and workflow configurations rather than establishing a general ranking of the underlying model families.

This distinction is clinically important because an apparently correct final diagnosis may arise from an unrealistic information advantage. Constrained patient simulation reduces this advantage by withholding facts until the system asks an appropriate question. It also creates measurable failure modes that static benchmarks cannot capture, including failure to ask about pregnancy, allergies, red flags, medication exposure, or symptom chronology.

The results further support multidimensional reporting. AI Doctor improved several reported outcome domains, substantially increased dialogue length, and reduced the number of cases outside the steps soft limit. A single average score would conceal this pattern; diagnosis, differential diagnosis, treatment, safety, and dialogue efficiency should therefore remain visible as separate outcomes when the underlying data are available.

The benchmark may also support training and assessment of clinicians or students because the same case structure can record history-taking coverage, diagnostic reasoning, and management recommendations. However, educational use would require separate validation, scoring standards appropriate to the learner level, and safeguards against using automated scores as the sole basis for high-stakes assessment.

\subsection{Strengths and Limitations}
\label{subsec:limitations}

The principal strengths of the framework are its end-to-end interactive design, explicit separation of the benchmark from the evaluated system, structured case-level reference fields, support for attachments, and an evaluation design that links scores to dialogue traces and evidence. The constrained patient simulator makes performance depend on targeted questioning rather than accidental disclosure of the complete case.

Several limitations must be considered. First, the archived export did not preserve a complete standardized metadata table, the full paired case-level output matrix, or all metric-specific denominators. Consequently, numerical case-mix reporting, subgroup analyses, and reconstruction of inferential statistics were not possible, and only outcomes that could be reconciled to the 254-case analysis are reported. Second, the synthetic case bank, evaluated system, scoring framework, and internal physician review were all associated with the same organization. The physician review was non-blinded and non-independent, creating a material risk of design, confirmation, and evaluator bias. Independent external cases and blinded adjudication are required to assess generalizability.

Third, the archived materials did not preserve all versioning elements required for exact reconstruction, including the freeze date, cryptographic hashes, complete provider-side model identifiers, all inference parameters, the complete paired case-level output matrix, and the original statistical analysis code. Consequently, the present comparison is descriptive and does not report inferential statistics.

Fourth, the patient simulator is a language-model proxy and cannot reproduce physical examination, non-verbal behavior, real patient uncertainty, or the full variability of clinical communication. Fifth, the automated evaluator has not undergone formal independent validation against blinded external physician adjudication of the same model outputs. Internal physician review of the cases and scoring framework does not substitute for independent evaluator validation. Semantic extraction errors and incomplete reference dictionaries may therefore affect benchmark scores.

Sixth, the GPT-5 basic reference was intentionally minimal and was not matched to Doctorina for output schema, clinical prompting, dialogue objectives, stopping conditions, attachment tooling, or agentic orchestration. The comparison therefore measures a full-system contrast, cannot isolate the contribution of individual Doctorina components, and does not establish general superiority over GPT-5. Finally, benchmark performance is not evidence of prospective clinical effectiveness, patient benefit, regulatory compliance, or safe autonomous deployment.

\section{Conclusions}
\label{sec:conclusions}

Doctorina MedBench reframes medical AI evaluation as a sequential, case-grounded clinical interaction rather than a static question-answering task. In the 254-case descriptive comparison, AI Doctor achieved higher values for Diagnosis Accuracy, Differential Accuracy, Treatment Accuracy, and Critical Conditions Passed than the deliberately restricted GPT-5 basic prompt-only reference. AI Doctor also conducted substantially longer dialogues and had fewer cases outside the steps soft limit. Because the comparator was not workflow matched, the internal physician review was non-blinded and non-independent, and the evaluator was not externally validated, the findings do not establish general superiority over GPT-5, clinical effectiveness, clinical safety, or readiness for real-world deployment. The resource is intended to support transparent comparative research and structured investigation of model behavior.

\vspace{6pt}

\supplementary{The following supporting information accompanies this manuscript: Supplementary File S1, Benchmark Data Dictionary and Case Schema; Supplementary File S2, Archived Prompt and Configuration Disclosure; and Supplementary File S3, Scoring Rules and Reproducibility Notes. Supplementary File S2 clearly distinguishes the verbatim GPT-5 basic reference instruction from functional constraints reconstructed from the archived documentation; the exact archived patient-simulator and evaluator prompt strings were not preserved. No synthetic attachments are included in the supplementary package.}

\authorcontributions{Conceptualization, A.K. and S.S.; methodology, A.K., S.S., P.S. and H.P.; software, S.S., P.S. and S.P.; validation, A.K., H.P. and P.S.; formal analysis, S.S. and P.S.; investigation, A.K. and H.P.; resources, A.K. and S.P.; data curation, A.K., H.P. and P.S.; writing---original draft preparation, A.K.; writing---review and editing, all authors; visualization, P.S. and S.S.; supervision, A.K. and S.P.; project administration, A.K.; funding acquisition, S.P. All authors have read and agreed to the submitted version of the manuscript and accept accountability for their contributions.}

\funding{This research received no external funding. The development of the benchmark, software, evaluation infrastructure, analysis, and manuscript was supported internally by A.I. Doctor Medical Assist LTD, Cyprus. The company, through the authors' employment and project roles, participated in study design, synthetic case development, software implementation, analysis, manuscript preparation, and the decision to submit the work for publication.}

\institutionalreview{Not applicable. All cases in the frozen benchmark were physician-authored synthetic scenarios. The study did not involve human participants, patient records, consultation transcripts, consultation logs from prototype testing, production-user interactions, human tissues, or identifiable or de-identified human data.}

\informedconsent{Not applicable. No human participants, patient-derived data, consultation logs from prototype testing, or production-user interactions were used in the benchmark-development or model-evaluation study.}

\dataavailability{A public non-commercial research release is available at \url{https://github.com/DoctorinaAI/doctorina-medbench-paper}. The field-level data dictionary, archived prompt/configuration disclosure, scoring specification, and reconciled aggregate results are provided in Supplementary Files S1--S3. The complete set of 261 attempted synthetic case records, individual model outputs, complete paired case-level export, exact patient-simulator and evaluator prompts, full runtime configuration, and executable scoring code are not included in the public supplementary package and were not fully preserved in the manuscript archive. The released materials support verification of the reported definitions, scoring logic, and aggregate counts, but they do not permit exact rerunning of the original evaluation. No human, patient-derived, prototype consultation-log, or production-interaction data are contained in the analyzed benchmark.}

\acknowledgments{The authors acknowledge the internal physicians and technical contributors who participated in synthetic case development, non-independent clinical review, software implementation, and quality assurance.}

\conflictsofinterest{All authors are affiliated with A.I. Doctor Medical Assist LTD, the developer of the Doctorina system evaluated in this study. The company supported synthetic benchmark development, internal physician review, software implementation, evaluation infrastructure, analysis, and manuscript preparation. To reduce---but not eliminate---the resulting risk of bias, the benchmark specification, frozen case set, parsing rules, and scoring logic were defined separately from the evaluated outputs; unfavorable as well as favorable outcome domains are reported; the GPT-5 comparison is explicitly limited to an architecture-level descriptive contrast; and the manuscript identifies the need for independent external case development, workflow-matched comparators, blinded evaluator validation, and replication by investigators not affiliated with the company.}

\abbreviations{Abbreviations}{
The following abbreviations are used in this manuscript:
\\

\noindent
\begin{tabular}{@{}ll}
AI & Artificial intelligence\\
D.O.T.S. & Diagnosis, Observations/Investigations, Treatment, and Step Count\\
LLM & Large language model\\
MRR & Mean reciprocal rank\\
\end{tabular}
}

\reftitle{References}

\isAPAandChicago{}{%

}

\isChicagoStyle{}{ }

\isAPAStyle{}{ }

\PublishersNote{}

\end{document}


\maketitle

\section*{Purpose and scope}
This supplement documents the public release schema used to represent the 254 synthetic cases released with Doctorina MedBench. It describes the fields present in the public JSONL file and the corresponding validation rules. The public data release is a reduced, metadata-limited case release and does not include the complete internal benchmark schema, complete paired model outputs, run-level audit records, complete category metadata, or executable scoring artifacts.

The archived evaluation attempted 261 synthetic case identifiers once with each system configuration. The paired complete-case comparative analysis retained 254 synthetic cases after excluding seven case identifiers with incomplete output in at least one configuration: 24, 47, 78, 96, 123, 220, and 239. The archived manuscript materials did not preserve a complete standardized metadata table or the full paired case-level export. Therefore, the public schema should be interpreted as the released case-record schema, not as a complete internal benchmark database schema.

\section*{Cardinality notation}
\begin{tabular}{ll}
\toprule
Notation & Meaning\\
\midrule
1 & Field is present in every public JSONL record\\
0..1 & Optional single value\\
0..* & Optional list\\
\bottomrule
\end{tabular}

\section*{Core public case schema}
\footnotesize
\setlength{\tabcolsep}{3pt}
\renewcommand{\arraystretch}{1.12}
\begin{longtable}{>{\raggedright\arraybackslash}p{0.24\linewidth}>{\raggedright\arraybackslash}p{0.16\linewidth}>{\centering\arraybackslash}p{0.12\linewidth}>{\raggedright\arraybackslash}p{0.40\linewidth}}
\caption{Field-level data dictionary for a released synthetic case record.}\\
\multicolumn{4}{p{0.94\linewidth}}{\emph{Note.} The table describes the public JSONL release schema. It does not describe a complete internal benchmark database or run-level audit export. Long machine-readable field identifiers are allowed to wrap at underscores to preserve the printed layout without changing field names.}\\[3pt]
\toprule
\textbf{Field} & \textbf{Type} & \textbf{Cardinality} & \textbf{Definition and validation rule}\\
\midrule
\endfirsthead
\toprule
\textbf{Field} & \textbf{Type} & \textbf{Cardinality} & \textbf{Definition and validation rule}\\
\midrule
\endhead

\field{id} & integer & 1 & Original synthetic case identifier. Must be unique within the public release.\\

\field{diagnosis} & string or null & 1 & Principal reference diagnosis label for the synthetic case. A null value indicates that the archived public export did not preserve a non-null value for this field.\\

\field{differential} & list[string] & 0..* & Reference alternative diagnostic hypotheses used for the released differential-diagnosis field.\\

\field{num\_\allowbreak{}steps} & integer or null & 1 & Reference dialogue-step target used as a soft behavioral audit. Null indicates that the archived public export did not preserve a numeric value for this field.\\

\field{intro} & string or null & 1 & Initial synthetic patient presentation shown to the evaluated system at the start of the dialogue.\\

\field{additional\_\allowbreak{}answers} & string or null & 1 & Case-grounded additional patient information that may be disclosed during the simulated dialogue.\\

\field{control\_\allowbreak{}questions} & string or null & 0..1 & Clinically relevant history-taking concepts or questions expected for the case, when preserved in the public export. Exact wording and order are not prescribed.\\

\field{weighted\_\allowbreak{}treatments\_\allowbreak{}new} & object, array, string, or null & 0..1 & Structured treatment-reference information when parseable, otherwise the original archived string value. Null indicates that no parseable or preserved value is available in the public export.\\

\field{treatment} & string or null & 0..1 & Treatment or management reference text preserved in the public export.\\

\field{critical\_\allowbreak{}conditions} & string or null & 0..1 & Safety-critical treatment or management conditions preserved for the case, when available.\\

\field{attachments} & list[string] & 0..* & Public attachment filenames associated with the synthetic case, when available. Attachment files are stored under \texttt{data/attachments/} and named \texttt{\{case\_id\}.\{ext\}}. The released attachments are synthetic or purpose-created and do not contain real patient files or patient-derived content.\\

\bottomrule
\end{longtable}
\normalsize

\section*{Synthetic-data provenance}
All cases covered by the manuscript are physician-authored synthetic scenarios. They were not derived from patient records, consultation transcripts, consultation logs from prototype testing, production-user interactions, or identifiable or de-identified human data. Any attachments referenced in the analyzed benchmark were synthetic or purpose-created and were not linked to patient records, prototype users, or production interactions.


\maketitle

\section*{Scope}
This supplement reports the prompt and configuration materials available for the 254-case paired complete-case analysis. The GPT-5 basic reference prompt, patient-simulator prompt template, evaluator prompt templates, and AI Doctor workflow instruction are reproduced below. Templates contain placeholders that were populated with case-specific runtime values. The exact runtime-expanded prompts for each individual case, provider-side model snapshot identifiers, sampling parameters, random seeds, run timestamps, full AI Doctor routing configuration, and software or analysis commit hashes were not preserved in the manuscript archive.

The comparative analysis used 254 paired complete cases from 261 attempted synthetic case identifiers; case identifiers 24, 47, 78, 96, 123, 220, and 239 were excluded because at least one configuration did not complete the run.

The prompt and configuration materials relate only to the physician-authored synthetic benchmark analyzed in the manuscript. The reported evaluation did not use patient records, consultation transcripts, consultation logs from prototype testing, production-user interactions, or identifiable or de-identified human data. Case-specific placeholders were populated from synthetic case records rather than from patient-derived materials.

\section*{Verbatim GPT-5 basic reference prompt}
\begin{lstlisting}[style=promptstyle]
Imagine that you are a doctor. Make a diagnosis or diagnoses.
\end{lstlisting}

This reference configuration intentionally omitted an explicit history-taking protocol, domain-specific routing, stopping policy, structured output schema, and equivalent agentic tooling. It is therefore an architecture-level zero-scaffold reference, not the strongest achievable GPT-5 workflow.

\section*{Archived patient-simulator prompt template}
Runtime values were inserted into placeholders such as \texttt{\{today\_datetime\}}, \texttt{\{patient\_intro\}}, and \texttt{\{potential\_answers\}} for each case.

\begin{lstlisting}[style=promptstyle]
PATIENT_SYSTEM_PROMPT = """
You are acting as a patient chatting with a doctor. Today is {today_datetime}.
**Your initial message to the doctor was:**
"{patient_intro}"

Your task is to answer the doctor's questions by selecting the most appropriate response from the following list:
{potential_answers}.

**Instructions:**
- **Directly Answer the Doctor's Last Question:** Only respond to the last question asked.
- **If the doctor's question is not covered by the responses above, reply with a negative response appropriate to the question in the same language as your {potential_answers}. Use phrases like "No, I don't...", "No, I haven't...", or "I'm not...", or "I don't know..." etc.**
- **Do Not End the Conversation Until the Doctor Stops Asking Questions:** Only conclude the conversation after the doctor finishes asking questions. If the doctor asks a follow-up question, answer it before saying goodbye.
- **For time-related questions** (e.g., "How long have you had this symptom?"): If no exact time frame is available in the responses, politely reply with an approximation or mention you don't recall.
- **If the same question is repeated:** Reformulate your previous answer, acknowledging that the question has already been answered. Example: "As I mentioned earlier ..." etc.
- **Do Not Reveal Any Symptoms Unless Specifically Asked:** Only share information related to the symptoms if the doctor explicitly asks about them. **DO NOT** volunteer additional details about your condition unless prompted.
- **Do Not Add Information:** Do not provide any information beyond the responses in the list or the specified default reply.
- **Be Concise:** Aim for answers between 10-20 words. Acknowledging a repeated question might require slightly more words.
- **Always use the same language for your list of answers given before.**

Imagine it's an online chat with a doctor."""
\end{lstlisting}

\section*{Archived evaluator prompt templates}
The evaluator used multiple task-specific prompt templates rather than a single evaluator prompt. Runtime values were inserted into placeholders such as \texttt{\{conversation\}}, \texttt{\{diagnosis\}}, \texttt{\{differential\}}, \texttt{\{should\_list\}}, \texttt{\{can\_list\}}, \texttt{\{questions\}}, \texttt{\{correct\_treatment\}}, \texttt{\{critical\_conditions\}}, \texttt{\{treatment\_conditions\}} and \texttt{\{correct\_diagnostic\_workup\}}.

\subsection*{EVAL\_PROMPT}
\begin{lstlisting}[style=promptstyle]
EVAL_PROMPT = """
You are a medical expert. Your job is to evaluate a conversation between a patient and an AI doctor assistant.

Your tasks:
1. Evaluate the accuracy and appropriateness of the AI doctor assistant's final diagnosis and differential diagnoses, considering only the final recommendation or summary at the end of the conversation.
2. Assess the quality of the clinical reasoning shown by the AI doctor assistant.
3. If the diagnosis or differential diagnoses are in a different language than the reference, translate as needed and compare their meanings.
4. Judge correctness based on clinical meaning, not strict wording--consider synonyms, equivalent medical terms, translations, and intent.
5. The diagnosis is correct if it matches the given correct diagnosis by meaning (including synonyms, translations, or clinically accepted alternatives).
6. Extract the proposed diagnosis strictly from the AI doctor assistant's final recommendation. It is forbidden to use any intermediate doctor thoughts. If there are multiple diagnoses in the final recommendation, extract all of them.

Instructions:
- Be precise and strictly follow these instructions.
- Be concise and to the point.
- Strictly adhere to the required output format.
- If the AI doctor assistant's final recommendation is not clear or absent, proposed diagnosis is "No diagnosis" with evidence "No recommendation".
- Always provide evidence for your answers. Evidence must be a direct quote from the conversation or "no evidence" if the information is not present.

Conversation:
{conversation}

Correct diagnosis:
{diagnosis}

Scope of differential diagnoses to evaluate:
{differential}
"""
\end{lstlisting}

\subsection*{DIAGNOSTIC\_EVALUATION\_PROMPT}
\begin{lstlisting}[style=promptstyle]
DIAGNOSTIC_EVALUATION_PROMPT = """
You are a medical expert. Review the conversation and indicate whether each listed diagnostic test WAS recommended/mentioned by the assistant doctor.

Conversation:
{conversation}

Mark each item as true if the assistant explicitly recommended or mentioned it, otherwise false.

Should be recommended (must be true to get full score):
{should_list}

Can be recommended (optional, no penalty if false):
{can_list}

Also list any ADDITIONAL diagnostic tests that the assistant recommended which are NOT in the two lists above as "unexpected_items".

Return ONLY a JSON with exactly this structure:
{{
  "should_be_recommended": {{"<item name>": <true/false>, ...}},
  "can_be_recommended": {{"<item name>": <true/false>, ...}},
  "unexpected_items": ["<test1>", "<test2>"]
}}
"""
\end{lstlisting}

\subsection*{HISTORY\_TAKING\_PROMPT}
\begin{lstlisting}[style=promptstyle]
HISTORY_TAKING_PROMPT = """
You are a highly skilled medical assistant tasked with evaluating a conversation between a user and an AI assistant.
Your job is to determine if the AI assistant successfully gathered all the required information from the user and asked the necessary questions.

Task:
- You will be given a conversation between a user and an AI assistant.
- Additionally, check if the AI assistant asked the necessary questions to gather the required information. Respond with either
  - 'True' if the AI assistant asked the necessary question. Add a quote from the conversation to justify your answer.
  - 'False' if the AI assistant did not ask the necessary question.

Additional Guidelines:
- Be precise and strictly follow the questions to be asked.
- Extract direct quotes from the conversation to justify any results. Always provide supporting evidence.

Questions to be asked:
{questions}

Conversation:
{conversation}
"""
\end{lstlisting}

\subsection*{TREATMENT\_CONDITIONS\_PROMPT}
\begin{lstlisting}[style=promptstyle]
TREATMENT_CONDITIONS_PROMPT = """
You are a highly skilled medical expert specializing in treatment validation. Your task is to evaluate prescribed treatments against a set of medical conditions to ensure patient safety and effectiveness.

General Approach:
1. Treatment Matching: Compare the treatment discussed in the conversation against the correct treatment provided.
   - If the treatment is mentioned in the conversation and in the correct treatment, it is a correct treatment.
   - If the treatment is not mentioned in the conversation but is present in the correct treatment, it is a missing treatment.
   - If the treatment is mentioned in the conversation absent in the correct treatment, it is a extra treatment.
   - If the treatment is mentioned in the conversation and in the correct treatment but with different details, it is a different treatment.
   - If the correct treatment provides alternative medications (e.g., "medicine A or B") and the conversation mentions either option, treat it as a correct treatment.

2. Critical Conditions Evaluation: Begin by checking all critical conditions. Each must be met for the treatment to be considered valid.
   - For each critical condition, determine:
      - FAILED: If a critical condition is explicitly violated in the conversation, mark as FAILED and provide a direct quote as evidence.
      - OK: If the conversation does not contain any violations of the critical condition.
      - NOT_APPLICABLE: If the critical condition is irrelevant to the specific patient case.

3. Treatment Conditions Evaluation: Evaluate the treatment from the conversation against the general treatment conditions.
   - For each treatment condition, determine:
     - OK:
        - If the condition is met.
        - If the correct condition specifies a single medication, but the conversation lists two or more medications as alternatives (using "or"), this is considered a partially correct treatment with excessive options.
     - FAILED:
       - If the condition is violated, provide a direct quote from the conversation as evidence.
     - NOT_APPLICABLE: If the condition does not apply to the patient.

4. Conditions for Inclusion in Missing Treatments:
    - A treatment or medication is added to missing treatments if any of the following applies:
        - It is explicitly listed in the correct treatment but is not mentioned at all in the conversation.
        - The correct treatment requires selection of one medication from a defined Group (e.g., Group 1, Group 2), and no medication from that group is mentioned in the conversation.
        - In group-based requirements, if only medications not belonging to the defined group are mentioned, the group is still considered missing.
        - The treatment is conditionally required (e.g., based on patient status or test results), and those conditions are met, but the corresponding treatment is still not provided in the conversation.

Additional Guidelines:
- Be precise and strictly adhere to the critical conditions.
- Extract direct quotes from the conversation to justify any results. Always provide supporting evidence.

Conversation:
{conversation}

Correct treatment:
{correct_treatment}

Critical conditions:
{critical_conditions}

Treatment conditions:
{treatment_conditions}
"""
\end{lstlisting}

\subsection*{DIAGNOSTIC\_CONDITIONS\_PROMPT}
\begin{lstlisting}[style=promptstyle]
DIAGNOSTIC_CONDITIONS_PROMPT = """
You are a highly skilled medical expert specializing in diagnostic evaluation. Your task is to extract diagnostic tests, laboratory tests, and imaging studies mentioned in the conversation and evaluate them against the expected diagnostic workup.

Instructions:
1. Extract all diagnostic tests, laboratory tests, and imaging studies mentioned in the conversation.
2. Identify which tests are recommended and which are mentioned but not recommended.
3. Compare against the expected diagnostic workup to identify matches and discrepancies.
4. Provide evidence from the conversation for each diagnostic test mentioned.

Guidelines:
- Focus on explicit mentions of diagnostic tests, lab work, and imaging
- Include both recommended tests and tests that are discussed but not recommended
- Provide direct quotes from the conversation as evidence
- Be precise in identifying the specific tests mentioned

Conversation:
{conversation}

Expected diagnostic workup:
{correct_diagnostic_workup}
"""
\end{lstlisting}

\section*{AI Doctor workflow instruction}
\begin{lstlisting}[style=promptstyle]
The AI Doctor conducts an iterative clinical dialogue with the user. It collects the presenting complaint, asks focused history questions, screens for urgent or red-flag symptoms, and requests relevant diagnostic information or attachments when available. During the conversation, it maintains a working clinical assessment and differential diagnosis, recommends appropriate investigations when indicated, and, once sufficient information has been gathered, provides treatment and management guidance together with follow-up and safety advice.

The workflow is designed to avoid fabricating unavailable information, to distinguish urgent from non-urgent presentations, and to keep clinical reasoning, diagnostic workup, treatment recommendations, follow-up instructions, and safety guidance logically separated in the final response.
\end{lstlisting}

\section*{Configuration disclosure and missing fields}
\small
\begin{longtable}{p{0.29\textwidth}p{0.22\textwidth}p{0.37\textwidth}}
\caption{Recoverability of configuration fields.}\\
\toprule
\textbf{Field} & \textbf{Status} & \textbf{Comment}\\
\midrule
\endfirsthead
\toprule
\textbf{Field} & \textbf{Status} & \textbf{Comment}\\
\midrule
\endhead
GPT-5 basic reference instruction & Preserved verbatim & Reproduced above.\\
Patient-simulator prompt template & Preserved as template & Runtime placeholders were populated with case-specific values.\\
Evaluator prompt templates & Preserved as templates & Multiple task-specific evaluator templates are reproduced above.\\
AI Doctor workflow instruction & Reproduced above & The high-level workflow instruction is provided for interpretability.\\
Provider/model family & Partially preserved & Described as GPT-5 for the reference; exact provider-side snapshot identifiers were not retained.\\
Reference temperature/top-p & Not preserved & Must not be inferred.\\
Reference random seed & Not preserved & Provider support and value are unknown.\\
Reference maximum output tokens & Not preserved & Must not be inferred.\\
Patient-simulator model and snapshot & Not preserved & The prompt template is available, but model metadata are unavailable.\\
Patient-simulator sampling parameters & Not preserved & Must not be inferred.\\
AI Doctor routing configuration & Not preserved in full & Only the workflow instruction and aggregate outputs are available in the manuscript archive.\\
AI Doctor component model snapshots & Not preserved & Must not be inferred.\\
Evaluator model and snapshot & Not preserved & Prompt templates are documented, but exact model metadata are unavailable.\\
Typed output schema & Partially preserved & Entity classes and evidence requirements are documented; exact serialized schema version is unavailable.\\
Software and analysis commit hashes & Not preserved & Exact reconstruction is not possible from the archived materials.\\
Run timestamps & Not preserved in the manuscript archive & Should be retained in future releases.\\
Runtime-expanded prompts for all 254 cases & Not preserved as a complete release artifact & Templates are provided, but the per-case expanded strings were not archived as a separate reproducibility package.\\
Paired case-level export & Not preserved in full & The comparative analysis is reported on 254 paired complete cases, with one completed output from each system per retained case; the complete record-level export was unavailable for exact rerunning.\\
\bottomrule
\end{longtable}
\normalsize

\section*{Interpretive consequence}
The comparison tests an AI Doctor workflow against a deliberately restricted GPT-5 basic prompt-only reference. It does not isolate the effect of individual AI Doctor components and does not establish superiority over a workflow-matched GPT-5 configuration. The prompt templates make the evaluation setup more inspectable, but missing model snapshots, inference parameters, seeds, routing metadata, runtime-expanded prompts, and complete case-level exports prevent exact rerunning of the archived experiment from the released materials alone.


\maketitle

\section*{Unit of reporting}
The archived evaluation attempted 261 synthetic case identifiers once with the GPT-5 basic prompt-only reference and once with AI Doctor. AI Doctor completed 255 of 261 attempted runs; incomplete AI Doctor case identifiers were 24, 47, 78, 96, 220, and 239. The GPT-5 basic prompt-only reference completed 260 of 261 attempted runs; the incomplete GPT-5 case identifier was 123. The paired complete-case analysis therefore retained 254 cases for which both configurations produced completed outputs. All reported values are descriptive. No p-values or confidence intervals are reported because the complete paired case-level output matrix and original analysis code were not preserved in the manuscript archive.

Incomplete runs were treated as unavailable comparative outputs rather than as incorrect clinical answers. For the agentic Doctorina workflow, non-completion may occur when the system does not force a final diagnosis or management plan after insufficient information has been obtained, or when further information or investigation would be required before finalizing recommendations. This behavior is distinct from a prompt-only response that produces an early diagnostic hypothesis after the initial presentation.

All scoring rules described in this supplement apply to physician-authored synthetic clinical scenarios. The analyzed benchmark did not include patient records, consultation transcripts, consultation logs from prototype testing, production-user interactions, or identifiable or de-identified human data. Consequently, the scoring specification should be interpreted as a benchmark-evaluation specification, not as a clinical audit of real patient encounters.

\section*{Paired-analysis inclusion rule}
\begin{table}[h]
\centering
\small
\begin{tabular}{lccc}
\toprule
\textbf{Case ID} & \textbf{AI Doctor completed} & \textbf{GPT-5 basic completed} & \textbf{Included in paired analysis} \\
\midrule
24 & No & Yes & No \\
47 & No & Yes & No \\
78 & No & Yes & No \\
96 & No & Yes & No \\
123 & Yes & No & No \\
220 & No & Yes & No \\
239 & No & Yes & No \\
\bottomrule
\end{tabular}
\end{table}

These seven attempted case identifiers were excluded from both systems' comparative outcome calculations. The retained denominator for all reported comparative metrics is therefore 254. No excluded run was imputed as a failed diagnosis or treatment recommendation.

\section*{Reconciled 254-case descriptive results}
\begin{table}[h]
\centering
\small
\begin{tabularx}{\textwidth}{>{\raggedright\arraybackslash}Xrr}
\toprule
\textbf{Metric} & \textbf{GPT-5 basic reference} & \textbf{AI Doctor} \\
\midrule
Diagnosis Accuracy & 86.61\% & 91.34\% \\
Differential Accuracy & 23.62\% & 46.69\% \\
Treatment Accuracy & 37.60\% & 54.46\% \\
Critical Conditions Passed & 98.82\% & 99.61\% \\
Total steps (doctor/patient pairs) & 167 & 3009 \\
Average steps (doctor/patient pairs) & 0.66 & 11.85 \\
Cases outside $\pm25\%$ of steps soft limit (num\_steps) & 242 & 140 \\
\bottomrule
\end{tabularx}
\end{table}

\section*{Diagnosis Accuracy}
Diagnosis Accuracy is 100 when at least one final diagnosis is judged correct and 0 otherwise.

\section*{Differential Accuracy}
Let $N_{\mathrm{correct\ differential}}$ be the number of reference differential diagnoses identified in the final differential list and $N_{\mathrm{reference\ differential}}$ the number in the case reference:
\[
D_{\mathrm{diff}}=100\times\frac{N_{\mathrm{correct\ differential}}}{N_{\mathrm{reference\ differential}}}.
\]
This is a legacy recall-like measure. It does not penalize every additional diagnosis and should not be interpreted as precision or ranking quality.

\section*{Treatment Accuracy}
Final treatment and management items are normalized into matching, extra, missing, or different categories:
\[
T=100\times\frac{N_{\mathrm{matching}}}{N_{\mathrm{matching}}+N_{\mathrm{extra}}+N_{\mathrm{missing}}+N_{\mathrm{different}}}.
\]
The denominator incorporates omissions and unsupported or discordant recommendations.

\section*{Critical Conditions Passed}
Each pre-specified safety-critical rule is assigned \texttt{OK}, \texttt{NOT\_APPLICABLE}, or failure. A case receives 100 only if every applicable critical condition is \texttt{OK} or \texttt{NOT\_APPLICABLE}; otherwise, it receives 0. Any reported percentage is the mean across analyzed cases.

\section*{Total steps (doctor/patient pairs), Average steps (doctor/patient pairs), and Cases outside $\pm25\%$ of steps soft limit (num\_steps)}
For $N_{\mathrm{assistant}}$ evaluated-system messages and $N_{\mathrm{user}}$ simulated-patient messages after the initial presentation:
\[
S=\min(N_{\mathrm{assistant}},N_{\mathrm{user}}).
\]
The initial patient presentation is excluded. The schema supports flagging a case outside the soft reference range when $S<0.75S_{\mathrm{target}}$ or $S>1.25S_{\mathrm{target}}$. This count is reported in the updated export as the number of cases outside the case-specific soft step range. Step count is interpreted jointly with completion and clinical outcomes.

\section*{Aggregation and missing outputs}
Metrics are reported as case-level means, counts, or totals across the 254-case paired complete-case analysis. Incomplete conversations remain in the relevant denominators and are scored from the output actually produced. No model output is imputed, and no case is removed because of poor performance. Because the complete paired output matrix and original analysis code were not retained in the manuscript archive, the available summaries do not support reproducible McNemar tests, Wilcoxon signed-rank tests, paired bootstrap confidence intervals, or subgroup inference.

\section*{Audit requirements for future releases}
A reproducible release should retain, for every evaluated case and system configuration, the raw dialogue, final response, structured extraction, evidence spans, normalized actions, item-level rule results, metric values, model and prompt versions, sampling parameters, random seed where supported, case and parent identifiers, attachment events, software commit hash, and analysis commit hash.